\documentclass{article}
\usepackage{spconf,amsmath,graphicx}
\usepackage[linesnumbered,ruled,vlined]{algorithm2e}
\usepackage{booktabs}
\usepackage{multirow}
\usepackage{longtable}
\usepackage{makecell}
\usepackage[table, svgnames, dvipsnames]{xcolor}
\usepackage[export]{adjustbox}
\usepackage[hyphens]{url}

\newcommand{\argmax}[1]{\underset{#1}{\operatorname{arg}\,\operatorname{max}}\;}

\usepackage{tikz,pgfplots}
\usepackage[caption=false,font=footnotesize]{subfig}

\definecolor{darkblue}{RGB}{57,79,99}
\definecolor{rosso}{RGB}{220,57,18}
\definecolor{giallo}{RGB}{255,153,0}
\definecolor{blu}{RGB}{102,140,217}
\definecolor{verde}{RGB}{16,150,24}
\definecolor{viola}{RGB}{153,0,153}
\definecolor{awesome}{rgb}{1.0, 0.13, 0.32}
\definecolor{ref}{rgb}{0.65,0.65,0.65} 
\definecolor{darkgreen}{RGB}{47,109,79}
\definecolor{darkblue}{RGB}{57,79,99}
\definecolor{rosso}{RGB}{220,57,18}
\definecolor{verde}{RGB}{16,150,24}
\definecolor{viola}{RGB}{153,0,153}
\definecolor{americanrose}{rgb}{1.0, 0.01, 0.24}
\definecolor{bostonuniversityred}{rgb}{0.8, 0.0, 0.0}
\definecolor{chocolate(traditional)}{rgb}{0.48, 0.25, 0.0}
\definecolor{violet(web)}{rgb}{0.93, 0.51, 0.93}
\definecolor{airforceblue}{rgb}{0.36, 0.54, 0.66}

\definecolor{almond}{rgb}{0.94, 0.87, 0.8}
\definecolor{amethyst}{rgb}{0.6, 0.4, 0.8}
\definecolor{bazaar}{rgb}{0.6, 0.47, 0.48}
\definecolor{britishracinggreen}{rgb}{0.0, 0.26, 0.15}
\definecolor{byzantine}{rgb}{0.74, 0.2, 0.64}
\definecolor{cadetblue}{rgb}{0.37, 0.62, 0.63}
\definecolor{cambridgeblue}{rgb}{0.64, 0.76, 0.68}
\definecolor{candypink}{rgb}{0.89, 0.44, 0.48}
\definecolor{caputmortuum}{rgb}{0.35, 0.15, 0.13}
\definecolor{cerulean}{rgb}{0.0, 0.48, 0.65}
\definecolor{corn}{rgb}{0.98, 0.93, 0.36}
\definecolor{darkbyzantium}{rgb}{0.36, 0.22, 0.33}
\definecolor{darkgoldenrod}{rgb}{0.72, 0.53, 0.04}
\definecolor{darkseagreen}{rgb}{0.56, 0.74, 0.56}
\definecolor{darkturquoise}{rgb}{0.0, 0.81, 0.82}
\definecolor{dimgray}{rgb}{0.41, 0.41, 0.41}
\definecolor{eggplant}{rgb}{0.38, 0.25, 0.32}

\definecolor{R50}{RGB}{175,0,3}
\definecolor{R18}{RGB}{233,107,34}
\definecolor{AN}{RGB}{38,181,66}

\title{EDGEFOOL: AN ADVERSARIAL IMAGE ENHANCEMENT FILTER}
\name{Ali Shahin Shamsabadi, Changjae Oh, Andrea Cavallaro\thanks{Andrea Cavallaro wishes to thank the Alan Turing Institute (EP/N510129/1), which is funded by the EPSRC, for its support through the project PRIMULA. Copyright 2020 IEEE. Published in the IEEE 2020 International Conference on Acoustics, Speech, and Signal Processing (ICASSP 2020), scheduled for 4-9 May, 2020, in Barcelona, Spain. Personal use of this material is permitted. However, permission to reprint/republish this material for advertising or promotional purposes or for creating new collective works for resale or redistribution to servers or lists, or to reuse any copyrighted component of this work in other works, must be obtained from the IEEE. Contact: Manager, Copyrights and Permissions / IEEE Service Center / 445 Hoes Lane / P.O. Box 1331 / Piscataway, NJ 08855-1331, USA. Telephone: + Intl. 908-562-3966.}}
\address{Centre for Intelligent Sensing, Queen Mary University of London, UK}

\begin{document}
\ninept
\maketitle
\begin{abstract}
Adversarial examples are intentionally perturbed images that mislead classifiers. These images can, however, be easily detected using denoising algorithms, when high-frequency spatial perturbations are used, or can be noticed by humans, when perturbations are large.
In this paper, we propose {EdgeFool}, an adversarial image enhancement filter that learns structure-aware adversarial perturbations. EdgeFool generates adversarial images with perturbations that enhance image details via training a fully convolutional neural network end-to-end with a multi-task loss function. This loss function accounts for both image detail enhancement and class misleading objectives. We evaluate {EdgeFool} on three classifiers (ResNet-50, ResNet-18 and AlexNet) using two datasets (ImageNet and Private-Places365) and compare it with six adversarial methods (DeepFool, SparseFool, Carlini-Wagner, SemanticAdv, Non-targeted and Private Fast Gradient Sign Methods). Code is available at \url{https://github.com/smartcameras/EdgeFool.git}.

\end{abstract}

\section{Introduction}
\label{sec:intro}

An adversarial image is generated by perturbing the pixel values of an original image to induce a Deep Neural Network (DNNs) to fail in its classification task. However, most adversarial images are \emph{detectable}~\cite{kurakin2016adversarial,Li2019,carlini2017towards} by defence mechanisms that use denoising algorithms, such as median filtering or bit-depth reduction~\cite{xu2017feature}, or, when large distortions are generated, the adversarial images are \emph{noticeable} to humans~\cite{hosseini2018semantic,modas2018sparsefool}. Moreover, most adversarial methods operate under a white-box attack assumption and need to access the parameters of a specific classifier. Therefore, the resulting perturbations do not generally {\em transfer} across classifiers~\cite{xie2018improving}.

Adversarial methods can be classified as constrained or unconstrained, depending on whether they generate bounded perturbations. {\em Constrained} adversarial methods~\cite{kurakin2016adversarial,Li2019,carlini2017towards,modas2018sparsefool,MoosaviDezfooli16} iteratively minimise an $\ell_p$ norm of the perturbations. Non-targeted Fast Gradient Sign Method (N-FGSM)~\cite{kurakin2016adversarial} and Private Fast Gradient Sign Method (P-FGSM)~\cite{Li2019} generate an adversarial perturbation whose $\ell_\infty$ norm is constrained. DeepFool~\cite{MoosaviDezfooli16} and Carlini-Wagner (CW)~\cite{carlini2017towards} constrain the $\ell_2$ difference between the original and adversarial image, while SparseFool~\cite{modas2018sparsefool} generates sparse adversarial perturbations based on the $\ell_1$ norm.  
However, these adversarial images are easily detectable~\cite{xu2017feature} and are not transferable~\cite{xie2018improving}. 

\begin{figure}[t]
    \centering
    \setlength\tabcolsep{1pt}
    \begin{tabular}{ccc}
        \includegraphics[valign=m,width=0.28\linewidth]{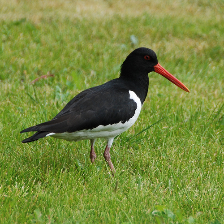}&
        \includegraphics[valign=m,width=0.28\linewidth]{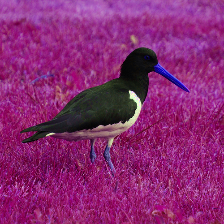}&
        \includegraphics[valign=m,width=0.28\linewidth]{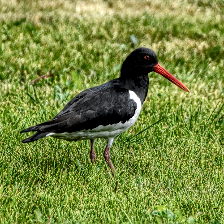}
         \\ 
         \includegraphics[valign=m,width=0.28\linewidth]{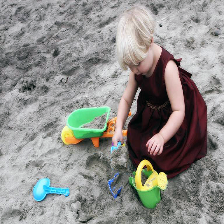}&
        \includegraphics[valign=m,width=0.28\linewidth]{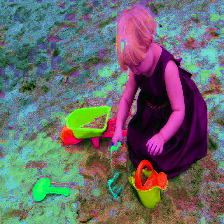}&
        \includegraphics[valign=m,width=0.28\linewidth]{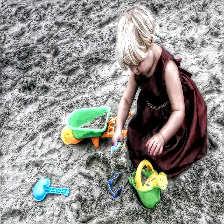}
        \\ 
         \textcolor{black}{(a)} & \textcolor{black}{(b)} & \textcolor{black}{(c)} 
    \end{tabular}
    \caption{Adversarial examples produced with {\em unconstrained} perturbations. (a) Original images. Adversarial images generated with (b)~SemanticAdv~\cite{hosseini2018semantic} and (c)~EdgeFool, the proposed method. Note the unnatural colours produced by SemanticAdv, and the enhanced details and natural colours produced by EdgeFool. }     
    \label{fig:advExample}
    
\end{figure}

More recent adversarial methods introduce {\em unconstrained} perturbations, for example by randomly shifting hue and saturation values~\cite{hosseini2018semantic}, by transferring new textures to input images~\cite{bhattad2019big} or by colourisation~\cite{bhattad2019big,zhu2019generating}. These adversarial methods are more transferable and less detectable than constrained adversarial methods, but unconstrained adversarial images may be severely degraded by unnatural  textures or colours (see Figure~\ref{fig:advExample}(b)). 

\begin{figure}[t!]
        \centering
        \includegraphics[width=0.5\textwidth]{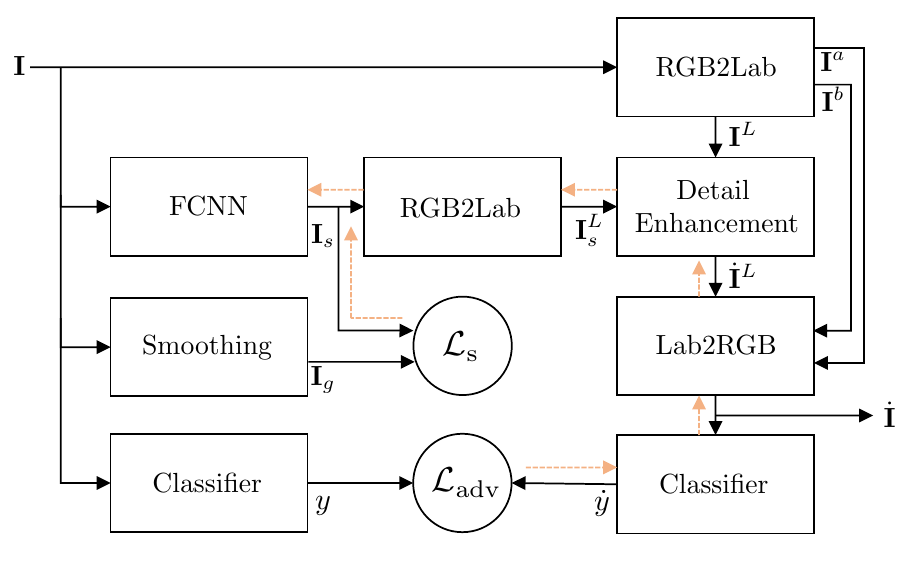}
        \caption{Block diagram of EdgeFool, which generates a detail-enhanced adversarial image, 
        $\dot{\mathbf{I}}$, by training a Fully Convolutional Neural Network (FCNN) using a smoothing loss, $\mathcal{L}_s$, and an adversarial loss, $\mathcal{L}_{\text{adv}}$ ({\protect\raisebox{2pt}{\protect\tikz \protect\draw[black,line width=0.1mm] (0,0) -- (0.3,0);}} forward pass;  {\protect\raisebox{2pt}{\protect\tikz \protect\draw[orange,line width=0.1mm,densely dashed] (0,0) -- (0.3,0);}} backward pass).
        The FCNN learns a smooth image, $\mathbf{I}_s$, based on the desired output, $\mathbf{I}_g$, of the image smoothing filter and the objective of  predicting a class of $\dot{\mathbf{I}}$, $\dot y$, that is different from  $y$, the class of the input image $\mathbf{I}$. }
        \label{fig:BlockDiagram}
\end{figure}

In this paper, we propose employing an image enhancement filter to generate adversarial images, with the objective of  reducing detectability and noticeability, and improving transferability.  We achieve these contrasting objectives with a structure-aware adversarial perturbation that enhances image details. The proposed adversarial image enhancement filter, {EdgeFool}, learns the adversarial perturbation by training a Fully Convolutional Neural Network (FCNN) end-to-end with a multi-task loss, which includes detail enhancement and class misleading objectives.  
Using image smoothing~\cite{xu2011image}, the network learns to decompose the image, isolating the details from smoother image structures. 
Then, the multi-task loss function guides the network to enhance details in a way that causes an incorrect classification. The block diagram of the proposed method is shown in Figure~\ref{fig:BlockDiagram}. 

We validate EdgeFool with object and scene classifiers on ImageNet~\cite{deng2009imagenet} and Private-Places365~\cite{zhou2017places} using deep residual neural networks with 50 layers (ResNet-50) and 18 layers (ResNet-18)~\cite{he2016deep}, and AlexNet~\cite{krizhevsky2012imagenet}. 
Experiments show that EdgeFool generates detail-enhanced adversarial images that are more transferable and less detectable than constrained adversarial methods.


\section{Proposed Method}
\label{sec:method}

\subsection{Preliminaries}

Adversarial methods aim to mislead a $D$-class classifier, $C(\cdot)$, by modifying the intensity values of an RGB image, $\mathbf{I}$, and generating an adversarial image, $\dot{\mathbf{I}}$, whose classification result, $C: \dot{\mathbf{I}} \rightarrow \dot{y}$, differs from that of the original image:
\begin{equation}
\dot{y}=C (\dot{\mathbf{I}})\neq y =C ({\mathbf{I}}),
\end{equation}
where $y$ is the predicted class of the original image.  

For each of the $D$ classes, the classifier $C(\cdot)$ predicts logit scores $\mathbf{z}=(z_{1},...,z_{i},...,z_{D})$ for $\mathbf{I}$, where  $z_{i} \in (-\infty,+\infty)$ is the logit score for class $i$. The logit scores are then normalised and the softmax is used to predict the probability $p_{i} \in [0,1]$ of each class $i$:
\begin{equation}
p_i = \frac{{e}^{z_i}}{\Sigma_{d=1}^D {e}^{z_d}},   
\end{equation}
where $\mathbf{p}=(p_{1},...,p_{i},...,p_{D})$ represents the probability of all the classes. 

The predicted class, $y$, is decided by a winner-take-all approach:
\begin{equation}
     y = \argmax{d=1,...,D}p_d.
\end{equation}
%

\subsection{Adversarial image enhancement}

\begin{figure}[t]
    \centering
    \setlength\tabcolsep{1pt}
    \begin{tabular}{cccc}
         \includegraphics[valign=m,width=0.25\linewidth]{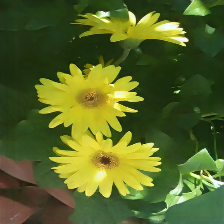}&
         \includegraphics[valign=m,width=0.25\linewidth]{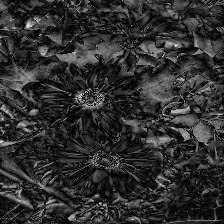}&
         \includegraphics[valign=m,width=0.25\linewidth]{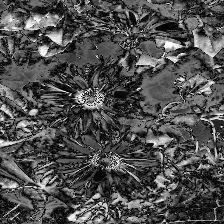} &
         \includegraphics[valign=m,width=0.25\linewidth]{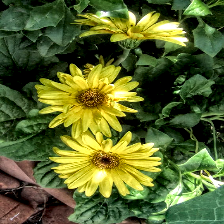}
         \\
        (a)   & (b)   & (c)  & (d) \\           
    \end{tabular}
    \caption{Intermediate results of the EdgeFool adversarial image generation process. (a) Image structure, $\mathbf{I}_s$. (b) Image  details, $\mathbf{I}_d^L$; (c) Enhanced details, $\dot{\mathbf{I}}^L-\mathbf{I}_s^L$. (d) Adversarial image, $\dot{\mathbf{I}}$. 
    Note that (b) and (c) are scaled for visualisation. }
      \label{fig:advLab}
\end{figure}

We propose an adversarial image enhancement filter, {EdgeFool}, whose  perturbations enhance details, preserve structure and maintain the original colours. EdgeFool decomposes $\mathbf{I}$ into its structural component, $\textbf{I}_s$ (Figure~\ref{fig:advLab}(a)), containing smooth regions, and a residual component, $\mathbf{I}_d$ (Figure~\ref{fig:advLab}(b)), corresponding to image details:
\begin{equation}
    \mathbf{I} = \mathbf{I}_s + \mathbf{I}_d. 
\end{equation}

We learn this image decomposition by training a FCNN~\cite{wu2018fast} with a multi-task loss, $\mathcal{L}$, which includes an image smoothing loss, $\mathcal{L}_\text{s}$, and the adversarial loss, $\mathcal{L}_\text{adv}$:
\begin{equation}
\label{eq:overall_loss}
\mathcal{L}=
\alpha \mathcal{L}_\text{s}\left(\mathbf{I}_s,\mathbf{I}_g \right) +
 \mathcal{L}_{\text{adv}}({\dot{\mathbf{I}}},\mathbf{I}),
\end{equation}
where the hyper-parameter $\alpha$ controls the relative importance of  $\mathcal{L}_\text{s}$ and $\mathcal{L}_\text{adv}$. 
The {\em smoothing loss}, $\mathcal{L}_\text{s}$, measures the difference between the learned structure and an image representation that excludes high spatial frequency components, $\mathbf{I}_g$, output by an $\ell_0$ structure-preserving smoothing filter~\cite{xu2011image}: 
\begin{equation}
\label{eq:loss_smooth}
\mathcal{L}_\text{s}\left(\mathbf{I}_s,\mathbf{I}_g \right)=
\|{ {\mathbf{I}_s-\mathbf{I}_g}}\|^2.
\end{equation}

We then avoid changing the colours of the input image and enhance the image details of the $L$ image channel only, after conversion to the $Lab$ colour space. The detail-enhanced adversarial image, $\dot{\mathbf{I}}$ (Figure~\ref{fig:advLab}(d)), is finally generated by  combining the channel with enhanced details, $\dot{\mathbf{I}}^L$, with the $a$ and $b$ colour channels of the original image, and then transforming the resulting image back to the RGB space:
\begin{equation}
\mathbf{\dot{I}} =
\begin{cases}
\dot{\mathbf{I}}^L = \Big(f\left(\frac{\mathbf{I}^L_s-v_1}{100},v_2\right) \cdot 100+v_1\Big) + f\left(\frac{\mathbf{I}^L_d}{100},v_3\right)\cdot 100,\\
\dot{\mathbf{I}}^a = \mathbf{I}^a ,\\    
\dot{\mathbf{I}}^b = \mathbf{I}^b,   
\end{cases}
\end{equation}
where $f(a,b)=\left(1+e^{-ab}\right)^{-1}-0.5$~\cite{fan2018image}, $v_1$ is a constant value that adjusts the midpoint of the sigmoid curve, $v_2$ and $v_3$ control the slope of the sigmoid curves in $\mathbf{I}^L_s$ and $\mathbf{I}^L_d$, respectively. The input to the sigmoid function is normalised by 100, the range of the $L$ channel. 

The {\em adversarial loss}, $\mathcal{L}_{\text{adv}}$, quantifies the difference between the score of the adversarial image $\dot{\mathbf{I}}$ belonging to the same class as ${\mathbf{I}}$, $\dot{z}_{y}$, and the maximum score among other classes~\cite{carlini2017towards}:  
\begin{equation}
\label{eq:loss_mislead}
\mathcal{L}_{\text{adv}}({\dot{\mathbf{I}}},\mathbf{I})= \dot{z}_{y} - \max\{{\dot{z}_i: i=1,...,D};i\neq y\}, 
\end{equation}
where $\dot{z}_i \in \dot{\mathbf{z}}$ is the logit score of $\dot{\mathbf{I}}$ for class $i$.

We iteratively train the whole pipeline end-to-end by minimising $\mathcal{L}$ in Eq.~\ref{eq:overall_loss} using the Adam optimiser~\cite{kingma2014adam}, until the generated $\dot{\mathbf{I}}$ misleads the target classifier and $\mathcal{L}_\text{s} < \tau$. 
We empirically set the threshold $\tau= 5\mathrm{e}^{-4}$ to make $\mathbf{I}_s$ close to $\mathbf{I}_g$, enabling the separation of the details for enhancement.

\begin{figure}[t!]
    \centering
    \setlength\tabcolsep{0.2pt}
    \begin{tabular}{cccc}
         \includegraphics[valign=m,width=0.25\linewidth]{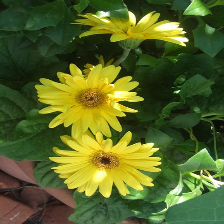}&&
         \includegraphics[valign=m,width=0.25\linewidth]{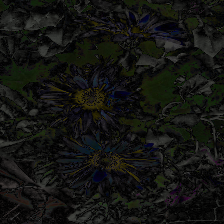}&
         \includegraphics[valign=m,width=0.25\linewidth]{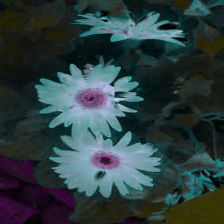}
         \\
          Original & & {EdgeFool} & SemanticAdv  \\
         \includegraphics[valign=m,width=0.25\linewidth]{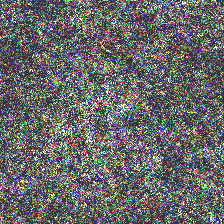}&
         \includegraphics[valign=m,width=0.25\linewidth]{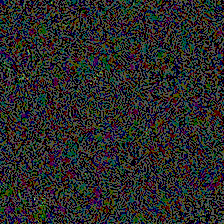}&
         \includegraphics[valign=m,width=0.25\linewidth]{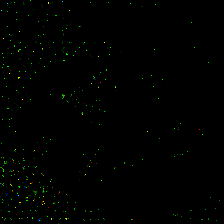}&
         \includegraphics[valign=m,width=0.25\linewidth]{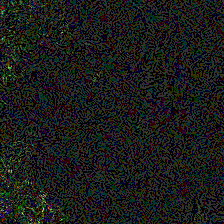}\\
         P-FGSM & DeepFool  & SparseFool &  CW \\
\vspace{-10pt}
    \end{tabular}
    \caption{Comparison of adversarial perturbations generated by different methods, namely EdgeFool, SemanticAdv~\cite{hosseini2018semantic}, P-FGSM~\cite{Li2019}, DeepFool~\cite{MoosaviDezfooli16}, SparseFool~\cite{modas2018sparsefool} and CW~\cite{carlini2017towards}. Note that the constrained perturbations (second row) are scaled for  visualisation.}
      \label{fig:advNoise}
      \vspace{-10pt}
\end{figure}

 Figure~\ref{fig:advNoise} shows an example of a perturbation generated by EdgeFool and compares it with the perturbations generated by state-of-the-art adversarial methods applied to the same image.

     \section{Validation}
\label{sec:val}

We compare the proposed method, {EdgeFool}, with six state-of-the-art adversarial methods: N-FGSM~\cite{kurakin2016adversarial}, P-FGSM~\cite{Li2019}, DeepFool~\cite{MoosaviDezfooli16}, SparseFool~\cite{modas2018sparsefool}, CW~\cite{carlini2017towards} and SemanticAdv~\cite{hosseini2018semantic}. We apply these adversarial methods to three state-of-the-art classifiers: ResNet with 50 layers (ResNet-50) and 18 layers (ResNet-18)~\cite{he2016deep}, and AlexNet~\cite{krizhevsky2012imagenet} on {Private-Places365}~\cite{zhou2017places,Mediaeval2018} and {ImageNet}~\cite{deng2009imagenet} as scene and object datasets. The Private-Places365 validation set includes 3K images of 60 privacy-sensitive scene classes. The ImageNet validation set includes 50 images of different size for 1K classes. In order to perform the evaluation on all 1K classes of ImageNet and reduce the computational cost, we consider 3K images by randomly selecting 3 images per class.

We used the PyTorch implementations provided by the authors for P-FGSM\footnote{{https://github.com/smartcameras/P-FGSM}}, DeepFool\footnote{{https://github.com/LTS4/DeepFool}} and SparseFool\footnote{{https://github.com/LTS4/SparseFool}} and we implemented N-FGSM. We also implemented SemanticAdv in PyTorch based on their Keras version\footnote{{https://github.com/HosseinHosseini/Semantic-Adversarial-Examples}}, while we used Foolbox\footnote{https://foolbox.readthedocs.io/en/stable/} for CW. We instantiate FCNN from the architecture implemented in~\cite{wu2018fast}. The input size is $224\times224\times3$. The FCNN architecture consists of 7 convolution layers with 24 intermediate feature maps and kernels of size $3\times3$.
The last convolution layer applies a $1\times1$ convolution that generates $\mathbf{I}_s$.
The dilation factors of each convolution layer are set to 1, 2, 4, 8, 16, 32, 1, and 1, respectively.
Leaky Rectified Linear Unit (L-ReLU) is applied after padding and normalising each intermediate convolutional layer, except the last convolution layer. For the detail enhancement, we follow the parameters in~\cite{fan2018image}, $v_1=56$, $v_2=1$, and $v_3=15$, which magnify the details by providing a steeper slope to the sigmoid curve for $\mathbf{I}^L_d$. Also, we choose $\alpha=10$ for the loss function. 

\pgfplotsset{ every non boxed x axis/.append style={x axis line style=-}}
\begin{figure}[t]
          \begin{tikzpicture}
           \node[inner sep=0pt] (whitehead) at (1.80,3)
          {\scriptsize Private-Places365};
          \node[inner sep=0pt] (whitehead) at (.27,2.75)
          {\tiny NF}; 
         \node[inner sep=0pt] (whitehead) at (.80,2.75)
          {\tiny PF}; 
         \node[inner sep=0pt] (whitehead) at (1.30,2.75)
          {\tiny DF}; 
        \node[inner sep=0pt] (whitehead) at (1.80,2.75)
          {\tiny SF};
        \node[inner sep=0pt] (whitehead) at (2.30,2.75)
          {\tiny CW};  
        \node[inner sep=0pt] (whitehead) at (2.77,2.75)
          {\tiny SA};  
        \node[inner sep=0pt] (whitehead) at (3.30,2.75)
          {\tiny \textbf{EF}};  
        \node[inner sep=0pt] (whitehead) at (.27,2.6)
          {\tiny \cite{kurakin2016adversarial}}; 
        \node[inner sep=0pt] (whitehead) at (.80,2.6)
          {\tiny \cite{Li2019}}; 
         \node[inner sep=0pt] (whitehead) at (1.30,2.6)
          {\tiny \cite{MoosaviDezfooli16}}; 
        \node[inner sep=0pt] (whitehead) at (1.80,2.6)
          {\tiny \cite{modas2018sparsefool}};
        \node[inner sep=0pt] (whitehead) at (2.30,2.6)
          {\tiny \cite{carlini2017towards}};  
        \node[inner sep=0pt] (whitehead) at (2.77,2.6)
          {\tiny \cite{hosseini2018semantic}};  
 
          \begin{axis}[
                  tiny,
                  axis lines=left, 
                  width=5cm,
                  height=4cm,
                  cycle list name=color list,
                  xticklabels={},
                  xtick=\empty,
                  xmin=0.00,
                  ymin=0.00,
                  ymax=1.01,
                  xmax=1.01,
                  smooth,
                  ylabel= Misleading rate,
                  enlarge x limits=0.01,
                  x axis line style={draw opacity=0},
                  y label style={font=\scriptsize, at={(axis description cs:.2,.5)},anchor=south},
              y tick label style={
                  /pgf/number format/.cd,
                      fixed,
                      fixed zerofill,
                      precision=1,
                  /tikz/.cd
              },
              x tick label style={
                  /pgf/number format/.cd,
                      fixed,
                      fixed zerofill,
                      precision=1,
                  /tikz/.cd}
                  ]
\addplot[
    scatter/classes={b=R50,v=R18, g=AN },
    scatter,
    black,
    mark=*,
    mark size=0.5mm,
    scatter src=explicit symbolic,
    ] table [meta=Class] {
         x          y     Class     
       .08	        1      b
       .05	       .2840   v
};\label{fig:FG_R50-R18}
\addplot[
    scatter/classes={b=R50,v=R18, g=AN },
    scatter,
    black,
    mark=*,
    mark size=0.5mm,
    scatter src=explicit symbolic,
    ] table [meta=Class] {
         x          y     Class     
       .08	        1      b
       .11	       .0727   g
};\label{fig:FG_R50-AN}
\addplot[
    scatter/classes={b=R50,v=R18, g=AN },
    scatter,
    black,
    mark=*,
    mark size=0.5mm,
    scatter src=explicit symbolic,
    ] table [meta=Class] {
         x          y     Class     
       .23	        1      b
       .20	       .301   v
};\label{fig:PFG_R50-R18}
\addplot[
    scatter/classes={b=R50,v=R18, g=AN },
    scatter,
    black,
    mark=*,
    mark size=0.5mm,
    scatter src=explicit symbolic,
    ] table [meta=Class] {
         x          y     Class     
       .23	        1      b
       .26	       .173   g
};\label{fig:PFG_R50-AN}
\addplot[
    scatter/classes={b=R50,v=R18, g=AN },
    scatter,
    black,
    mark=*,
    mark size=0.5mm,
    scatter src=explicit symbolic,
    ] table [meta=Class] {
         x          y     Class     
       .38	      .9573   b
       .35	      .1073   v
};\label{fig:DF_R50-R18}
\addplot[
    scatter/classes={b=R50,v=R18, g=AN },
    scatter,
    black,
    mark=*,
    mark size=0.5mm,
    scatter src=explicit symbolic,
    ] table [meta=Class] {
         x          y     Class     
       .38	       .9573      b
       .41	       .03        g
};\label{fig:DF_R50-AN}
\addplot[
    scatter/classes={b=R50,v=R18, g=AN },
    scatter,
    black,
    mark=*,
    mark size=0.5mm,
    scatter src=explicit symbolic,
    ] table [meta=Class] {
         x          y     Class     
       .53	      .9980   b
       .50	      .1510   v
};\label{fig:SF_R50-R18}
\addplot[
    scatter/classes={b=R50,v=R18, g=AN },
    scatter,
    black,
    mark=*,
    mark size=0.5mm,
    scatter src=explicit symbolic,
    ] table [meta=Class] {
         x          y     Class     
       .53	       .9980      b
       .56	       .1267        g
};\label{fig:SF_R50-AN}
\addplot[
    scatter/classes={b=R50,v=R18, g=AN },
    scatter,
    black,
    mark=*,
    mark size=0.5mm,
    scatter src=explicit symbolic,
    ] table [meta=Class] {
         x          y     Class     
       .68	      .9947   b
       .65	      .1010   v
};\label{fig:cw_R50-R18}
\addplot[
    scatter/classes={b=R50,v=R18, g=AN },
    scatter,
    black,
    mark=*,
    mark size=0.5mm,
    scatter src=explicit symbolic,
    ] table [meta=Class] {
         x          y     Class     
       .68	       .9947      b
       .71	       .0260      g
};\label{fig:cw_R50-AN}
\addplot[
    scatter/classes={b=R50,v=R18, g=AN },
    scatter,
    black,
    mark=*,
    mark size=0.5mm,
    scatter src=explicit symbolic,
    ] table [meta=Class] {
         x          y     Class     
       .83	      .9363   b
       .80	      .5633   v
};\label{fig:sa_R50-R18}
\addplot[
    scatter/classes={b=R50,v=R18, g=AN },
    scatter,
    black,
    mark=*,
    mark size=0.5mm,
    scatter src=explicit symbolic,
    ] table [meta=Class] {
         x          y     Class     
       .83	       .9363      b
       .86	       .7130      g
};\label{fig:sa_R50-AN}
\addplot[
    scatter/classes={b=R50,v=R18, g=AN },
    scatter,
    black,
    mark=*,
    mark size=0.5mm,
    scatter src=explicit symbolic,
    ] table [meta=Class] {
         x          y     Class     
       .98	      .9947   b
       .95	      .4940   v
};\label{fig:ef_R50-R18}
\addplot[
    scatter/classes={b=R50,v=R18, g=AN },
    scatter,
    black,
    mark=*,
    mark size=0.5mm,
    scatter src=explicit symbolic,
    ] table [meta=Class] {
         x          y     Class     
       .98	       .9947      b
       1.01	       .5240      g
};\label{fig:ef_R50-AN}

\end{axis}
\end{tikzpicture}
\begin{tikzpicture}
           \node[inner sep=0pt] (whitehead) at (1.80,3)
          {\scriptsize ImageNet};
          \node[inner sep=0pt] (whitehead) at (.27,2.75)
          {\tiny NF}; 
         \node[inner sep=0pt] (whitehead) at (.80,2.75)
          {\tiny PF}; 
         \node[inner sep=0pt] (whitehead) at (1.30,2.75)
          {\tiny DF}; 
        \node[inner sep=0pt] (whitehead) at (1.80,2.75)
          {\tiny SF};
        \node[inner sep=0pt] (whitehead) at (2.30,2.75)
          {\tiny CW};  
        \node[inner sep=0pt] (whitehead) at (2.77,2.75)
          {\tiny SA};  
        \node[inner sep=0pt] (whitehead) at (3.30,2.75)
          {\tiny \textbf{EF}};  
        \node[inner sep=0pt] (whitehead) at (.27,2.6)
          {\tiny \cite{kurakin2016adversarial}}; 
        \node[inner sep=0pt] (whitehead) at (.80,2.6)
          {\tiny \cite{Li2019}}; 
         \node[inner sep=0pt] (whitehead) at (1.30,2.6)
          {\tiny \cite{MoosaviDezfooli16}}; 
        \node[inner sep=0pt] (whitehead) at (1.80,2.6)
          {\tiny \cite{modas2018sparsefool}};
        \node[inner sep=0pt] (whitehead) at (2.30,2.6)
          {\tiny \cite{carlini2017towards}};  
        \node[inner sep=0pt] (whitehead) at (2.77,2.6)
          {\tiny \cite{hosseini2018semantic}};      
          \begin{axis}[
                  tiny,
                  axis lines=left, 
                  width=5cm,
                  height=4cm,
                  cycle list name=color list,
                  xticklabels={},
                  xtick=\empty,
                  xmin=0.00,
                  ymin=0.00,
                  ymax=1.01,
                  xmax=1.01,
                  smooth,
                  enlarge x limits=0.01,
                  x axis line style={draw opacity=0},
                  y label style={font=\tiny, at={(axis description cs:.2,.5)},anchor=south},
              y tick label style={
                  /pgf/number format/.cd,
                      fixed,
                      fixed zerofill,
                      precision=1,
                  /tikz/.cd
              },
              x tick label style={
                  /pgf/number format/.cd,
                      fixed,
                      fixed zerofill,
                      precision=1,
                  /tikz/.cd}
                  ]
\addplot[
    scatter/classes={b=R50,v=R18, g=AN },
    scatter,
    black,
    mark=*, mark size=0.5mm,
    mark size=0.5mm,
    scatter src=explicit symbolic,
    ] table [meta=Class] {
         x          y     Class     
       .08	       .8723      b
       .05	       .1230   v
};\label{fig:FG_R50-R18}
\addplot[
    scatter/classes={b=R50,v=R18, g=AN },
    scatter,
    black,
    mark=*, mark size=0.5mm,
    scatter src=explicit symbolic,
    ] table [meta=Class] {
         x          y     Class     
       .08	        .8723      b
       .11	       .0873   g
};\label{fig:FG_R50-AN}
\addplot[
    scatter/classes={b=R50,v=R18, g=AN },
    scatter,
    black,
    mark=*, mark size=0.5mm,
    scatter src=explicit symbolic,
    ] table [meta=Class] {
         x          y     Class     
       .23	        1      b
       .20	       .17   v
};\label{fig:PFG_R50-R18}
\addplot[
    scatter/classes={b=R50,v=R18, g=AN },
    scatter,
    black,
    mark=*, mark size=0.5mm,
    scatter src=explicit symbolic,
    ] table [meta=Class] {
         x          y     Class     
       .23	        1      b
       .26	       .1037   g
};\label{fig:PFG_R50-AN}
\addplot[
    scatter/classes={b=R50,v=R18, g=AN },
    scatter,
    black,
    mark=*, mark size=0.5mm,
    scatter src=explicit symbolic,
    ] table [meta=Class] {
         x          y     Class     
       .38	      .983   b
       .35	      .071   v
};\label{fig:DF_R50-R18}
\addplot[
    scatter/classes={b=R50,v=R18, g=AN },
    scatter,
    black,
    mark=*, mark size=0.5mm,
    scatter src=explicit symbolic,
    ] table [meta=Class] {
         x          y     Class     
       .38	       .983      b
       .41	       .018       g
};\label{fig:DF_R50-AN}
\addplot[
    scatter/classes={b=R50,v=R18, g=AN },
    scatter,
    black,
    mark=*, mark size=0.5mm,
    scatter src=explicit symbolic,
    ] table [meta=Class] {
         x          y     Class     
       .53	      .9870   b
       .50	      .1670   v
};\label{fig:SF_R50-R18}
\addplot[
    scatter/classes={b=R50,v=R18, g=AN },
    scatter,
    black,
    mark=*, mark size=0.5mm,
    scatter src=explicit symbolic,
    ] table [meta=Class] {
         x          y     Class     
       .53	       .9870      b
       .56	       .1757        g
};\label{fig:SF_R50-AN}
\addplot[
    scatter/classes={b=R50,v=R18, g=AN },
    scatter,
    black,
    mark=*, mark size=0.5mm,
    scatter src=explicit symbolic,
    ] table [meta=Class] {
         x          y     Class     
       .68	      .9907   b
       .65	      .054   v
};\label{fig:cw_R50-R18}
\addplot[
    scatter/classes={b=R50,v=R18, g=AN },
    scatter,
    black,
    mark=*, mark size=0.5mm,
    scatter src=explicit symbolic,
    ] table [meta=Class] {
         x          y     Class     
       .68	       .9907      b
       .71	       .0176      g
};\label{fig:cw_R50-AN}
\addplot[
    scatter/classes={b=R50,v=R18, g=AN },
    scatter,
    black,
    mark=*, mark size=0.5mm,
    scatter src=explicit symbolic,
    ] table [meta=Class] {
         x          y     Class     
       .83	      .8897   b
       .80	      .5400   v
};\label{fig:sa_R50-R18}
\addplot[
    scatter/classes={b=R50,v=R18, g=AN },
    scatter,
    black,
    mark=*, mark size=0.5mm,
    scatter src=explicit symbolic,
    ] table [meta=Class] {
         x          y     Class     
       .83	       .8897      b
       .86	       .7697      g
};\label{fig:sa_R50-AN}
\addplot[
    scatter/classes={b=R50,v=R18, g=AN },
    scatter,
    black,
    mark=*, mark size=0.5mm,
    scatter src=explicit symbolic,
    ] table [meta=Class] {
         x          y     Class     
       .98	      .9880   b
       .95	      .3607   v
};\label{fig:ef_R50-R18}
\addplot[
    scatter/classes={b=R50,v=R18, g=AN },
    scatter,
    black,
    mark=*, mark size=0.5mm,
    scatter src=explicit symbolic,
    ] table [meta=Class] {
         x          y     Class     
       .98	       .9880      b
       1.01	       .5093      g
};\label{fig:ef_R50-AN}
\end{axis}
\end{tikzpicture}\\
\begin{tikzpicture}
\node[inner sep=0pt] (whitehead) at (.27,2.6)
{\tiny NF}; 
\node[inner sep=0pt] (whitehead) at (.80,2.6)
{\tiny PF}; 
\node[inner sep=0pt] (whitehead) at (1.30,2.6)
{\tiny DF}; 
\node[inner sep=0pt] (whitehead) at (1.80,2.6)
{\tiny SF};
\node[inner sep=0pt] (whitehead) at (2.30,2.6)
{\tiny CW};  
\node[inner sep=0pt] (whitehead) at (2.77,2.6)
{\tiny SA};  
\node[inner sep=0pt] (whitehead) at (3.30,2.6)
{\tiny \textbf{EF}};    
\begin{axis}[
tiny,
axis lines=left, 
width=5cm,
height=4cm,
cycle list name=color list,
xticklabels={},
xtick=\empty,
xmin=0.00,
ymin=0.00,
ymax=1.01,
xmax=1.01,
smooth,
ylabel= Misleading rate,
enlarge x limits=0.01,
x axis line style={draw opacity=0},
y label style={font=\scriptsize, at={(axis description cs:.2,.5)},anchor=south},
y tick label style={
/pgf/number format/.cd,
fixed,
fixed zerofill,
precision=1,
/tikz/.cd
},
x tick label style={
/pgf/number format/.cd,
fixed,
fixed zerofill,
precision=1,
/tikz/.cd}
]
\addplot[
scatter/classes={b=R50,v=R18, g=AN },
scatter,
black,
mark=*, mark size=0.5mm,
scatter src=explicit symbolic,
] table [meta=Class] {
x          y     Class     
.08	        1      v
.05	       .2310   b
};\label{fig:FG_R18-R50}
\addplot[
scatter/classes={b=R50,v=R18, g=AN },
scatter,
black,
mark=*, mark size=0.5mm,
scatter src=explicit symbolic,
] table [meta=Class] {
x          y     Class     
.08	        1      v
.11	       .0810   g
};\label{fig:FG_R18-AN}
\addplot[
scatter/classes={b=R50,v=R18, g=AN },
scatter,
black,
mark=*, mark size=0.5mm,
scatter src=explicit symbolic,
] table [meta=Class] {
x          y     Class     
.23	        1      v
.20	       .2677   b
};\label{fig:PFG_R18-R50}
\addplot[
scatter/classes={b=R50,v=R18, g=AN },
scatter,
black,
mark=*, mark size=0.5mm,
scatter src=explicit symbolic,
] table [meta=Class] {
x          y     Class     
.23	        1      v
.26	       .185   g
};\label{fig:PFG_R18-AN}
\addplot[
scatter/classes={b=R50,v=R18, g=AN },
scatter,
black,
mark=*, mark size=0.5mm,
scatter src=explicit symbolic,
] table [meta=Class] {
x          y     Class     
.38	      .9690   v
.35	      .0090   b
};\label{fig:DF_R18-R50}
\addplot[
scatter/classes={b=R50,v=R18, g=AN },
scatter,
black,
mark=*, mark size=0.5mm,
scatter src=explicit symbolic,
] table [meta=Class] {
x          y     Class     
.38	       .9690      v
.41	       .0300        g
};\label{fig:DF_R18-AN}
\addplot[
scatter/classes={b=R50,v=R18, g=AN },
scatter,
black,
mark=*, mark size=0.5mm,
scatter src=explicit symbolic,
] table [meta=Class] {
x          y     Class     
.53	      .9987   v
.50	      .1007   b
};\label{fig:SF_R18-R50}
\addplot[
scatter/classes={b=R50,v=R18, g=AN },
scatter,
black,
mark=*, mark size=0.5mm,
scatter src=explicit symbolic,
] table [meta=Class] {
x          y     Class     
.53	       .9987     v
.56	       .1203        g
};\label{fig:SF_R18-AN}
\addplot[
scatter/classes={b=R50,v=R18, g=AN },
scatter,
black,
mark=*, mark size=0.5mm,
scatter src=explicit symbolic,
] table [meta=Class] {
x          y     Class     
.68	      .9960   v
.65	      .0840   b
};\label{fig:cw_R18-R50}
\addplot[
scatter/classes={b=R50,v=R18, g=AN },
scatter,
black,
mark=*, mark size=0.5mm,
scatter src=explicit symbolic,
] table [meta=Class] {
x          y     Class     
.68	       .9960      v
.71	       .0287      g
};\label{fig:cw_R18-AN}
\addplot[
scatter/classes={b=R50,v=R18, g=AN },
scatter,
black,
mark=*, mark size=0.5mm,
scatter src=explicit symbolic,
] table [meta=Class] {
x          y     Class     
.83	      .9540   v
.80	      .4803   b
};\label{fig:sa_R18-R50}
\addplot[
scatter/classes={b=R50,v=R18, g=AN },
scatter,
black,
mark=*, mark size=0.5mm,
scatter src=explicit symbolic,
] table [meta=Class] {
x          y     Class     
.83	       .9540      v
.86	       .7143      g
};\label{fig:sa_R18-AN}
\addplot[
scatter/classes={b=R50,v=R18, g=AN },
scatter,
black,
mark=*, mark size=0.5mm,
scatter src=explicit symbolic,
] table [meta=Class] {
x          y     Class     
.98	      .9906   v
.95	      .3864   b
};\label{fig:ef_R18-R50}
\addplot[
scatter/classes={b=R50,v=R18, g=AN },
scatter,
black,
mark=*, mark size=0.5mm,
scatter src=explicit symbolic,
] table [meta=Class] {
x          y     Class     
.98	       .9906      v
1.01	       .4877      g
};\label{fig:ef_R18-AN}
\end{axis}
\end{tikzpicture}
%
\begin{tikzpicture}
\node[inner sep=0pt] (whitehead) at (.27,2.6)
{\tiny NF}; 
\node[inner sep=0pt] (whitehead) at (.80,2.6)
{\tiny PF}; 
\node[inner sep=0pt] (whitehead) at (1.30,2.6)
{\tiny DF}; 
\node[inner sep=0pt] (whitehead) at (1.80,2.6)
{\tiny SF};
\node[inner sep=0pt] (whitehead) at (2.30,2.6)
{\tiny CW};  
\node[inner sep=0pt] (whitehead) at (2.77,2.6)
{\tiny SA};  
\node[inner sep=0pt] (whitehead) at (3.30,2.6)
{\tiny \textbf{EF}};    
\begin{axis}[
tiny,
axis lines=left, 
width=5cm,
height=4cm,
cycle list name=color list,
xticklabels={},
xtick=\empty,
xmin=0.00,
ymin=0.00,
ymax=1.01,
xmax=1.01,
smooth,
enlarge x limits=0.01,
x axis line style={draw opacity=0},
y label style={font=\tiny, at={(axis description cs:.2,.5)},anchor=south},
y tick label style={
/pgf/number format/.cd,
fixed,
fixed zerofill,
precision=1,
/tikz/.cd
},
x tick label style={
/pgf/number format/.cd,
fixed,
fixed zerofill,
precision=1,
/tikz/.cd}
]
\addplot[
scatter/classes={b=R50,v=R18, g=AN },
scatter,
black,
mark=*, mark size=0.5mm,
scatter src=explicit symbolic,
] table [meta=Class] {
x          y     Class     
.08	        .9450      v
.05	       .1433   b
};\label{fig:FG_R18-R50}
\addplot[
scatter/classes={b=R50,v=R18, g=AN },
scatter,
black,
mark=*, mark size=0.5mm,
scatter src=explicit symbolic,
] table [meta=Class] {
x          y     Class     
.08	        .9450      v
.11	       .0997   g
};\label{fig:FG_R18-AN}
\addplot[
scatter/classes={b=R50,v=R18, g=AN },
scatter,
black,
mark=*, mark size=0.5mm,
scatter src=explicit symbolic,
] table [meta=Class] {
x          y     Class     
.23	        1      v
.20	       .1400   b
};\label{fig:PFG_R18-R50}
\addplot[
scatter/classes={b=R50,v=R18, g=AN },
scatter,
black,
mark=*, mark size=0.5mm,
scatter src=explicit symbolic,
] table [meta=Class] {
x          y     Class     
.23	        1      v
.26	       .1037   g
};\label{fig:PFG_R18-AN}
\addplot[
scatter/classes={b=R50,v=R18, g=AN },
scatter,
black,
mark=*, mark size=0.5mm,
scatter src=explicit symbolic,
] table [meta=Class] {
x          y     Class     
.38	      .991   v
.35	      .055   b
};\label{fig:DF_R18-R50}
\addplot[
scatter/classes={b=R50,v=R18, g=AN },
scatter,
black,
mark=*, mark size=0.5mm,
scatter src=explicit symbolic,
] table [meta=Class] {
x          y     Class     
.38	       .991      v
.41	       .017        g
};\label{fig:DF_R18-AN}
\addplot[
scatter/classes={b=R50,v=R18, g=AN },
scatter,
black,
mark=*, mark size=0.5mm,
scatter src=explicit symbolic,
] table [meta=Class] {
x          y     Class     
.53	      .9970   v
.50	      .0857   b
};\label{fig:SF_R18-R50}
\addplot[
scatter/classes={b=R50,v=R18, g=AN },
scatter,
black,
mark=*, mark size=0.5mm,
scatter src=explicit symbolic,
] table [meta=Class] {
x          y     Class     
.53	       .9970      v
.56	       .1343        g
};\label{fig:SF_R18-AN}
\addplot[
scatter/classes={b=R50,v=R18, g=AN },
scatter,
black,
mark=*, mark size=0.5mm,
scatter src=explicit symbolic,
] table [meta=Class] {
x          y     Class     
.68	      .9936   v
.65	      .0426   b
};\label{fig:cw_R18-R50}
\addplot[
scatter/classes={b=R50,v=R18, g=AN },
scatter,
black,
mark=*, mark size=0.5mm,
scatter src=explicit symbolic,
] table [meta=Class] {
x          y     Class     
.68	       .9936      v
.71	       .0236      g
};\label{fig:cw_R18-AN}
\addplot[
scatter/classes={b=R50,v=R18, g=AN },
scatter,
black,
mark=*, mark size=0.5mm,
scatter src=explicit symbolic,
] table [meta=Class] {
x          y     Class     
.83	      .9307   v
.80	      .4217   b
};\label{fig:sa_R18-R50}
\addplot[
scatter/classes={b=R50,v=R18, g=AN },
scatter,
black,
mark=*, mark size=0.5mm,
scatter src=explicit symbolic,
] table [meta=Class] {
x          y     Class     
.83	       .9307      v
.86	       .7567      g
};\label{fig:sa_R18-AN}
\addplot[
scatter/classes={b=R50,v=R18, g=AN },
scatter,
black,
mark=*, mark size=0.5mm,
scatter src=explicit symbolic,
] table [meta=Class] {
x          y     Class     
.98	      .9880   v
.95	      .2737   b
};\label{fig:ef_R18-R50}
\addplot[
scatter/classes={b=R50,v=R18, g=AN },
scatter,
black,
mark=*, mark size=0.5mm,
scatter src=explicit symbolic,
] table [meta=Class] {
x          y     Class     
.98	       .9880      v
1.01	       .4890      g
};\label{fig:ef_R18-AN}
\end{axis}
\end{tikzpicture}\\
%
\begin{tikzpicture}
\node[inner sep=0pt] (whitehead) at (.27,2.6)
{\tiny NF}; 
\node[inner sep=0pt] (whitehead) at (.80,2.6)
{\tiny PF}; 
\node[inner sep=0pt] (whitehead) at (1.30,2.6)
{\tiny DF}; 
\node[inner sep=0pt] (whitehead) at (1.80,2.6)
{\tiny SF};
\node[inner sep=0pt] (whitehead) at (2.30,2.6)
{\tiny CW};  
\node[inner sep=0pt] (whitehead) at (2.77,2.6)
{\tiny SA};  
\node[inner sep=0pt] (whitehead) at (3.30,2.6)
{\tiny \textbf{EF}};    
\begin{axis}[
tiny,
axis lines=left, 
width=5cm,
height=4cm,
cycle list name=color list,
xticklabels={},
xtick=\empty,
xmin=0.00,
ymin=0.00,
ymax=1.01,
xmax=1.01,
smooth,
ylabel= Misleading rate,
enlarge x limits=0.01,
x axis line style={draw opacity=0},
y label style={font=\scriptsize, at={(axis description cs:.2,.5)},anchor=south},
y tick label style={
/pgf/number format/.cd,
fixed,
fixed zerofill,
precision=1,
/tikz/.cd
},
x tick label style={
/pgf/number format/.cd,
fixed,
fixed zerofill,
precision=1,
/tikz/.cd}
]
\addplot[
scatter/classes={b=R50,v=R18, g=AN },
scatter,
black,
mark=*, mark size=0.5mm,
scatter src=explicit symbolic,
] table [meta=Class] {
x          y     Class     
.08	        1      g
.05	       .0613   b
};\label{fig:FG_AN-R50}
\addplot[
scatter/classes={b=R50,v=R18, g=AN },
scatter,
black,
mark=*, mark size=0.5mm,
scatter src=explicit symbolic,
] table [meta=Class] {
x          y     Class     
.08	        1      g
.11	       .0807   v
};\label{fig:FG_AN-R18}
\addplot[
scatter/classes={b=R50,v=R18, g=AN },
scatter,
black,
mark=*, mark size=0.5mm,
scatter src=explicit symbolic,
] table [meta=Class] {
x          y     Class     
.23	        1      g
.20	       .1674   b
};\label{fig:PFG_AN-R50}
\addplot[
scatter/classes={b=R50,v=R18, g=AN },
scatter,
black,
mark=*, mark size=0.5mm,
scatter src=explicit symbolic,
] table [meta=Class] {
x          y     Class     
.23	        1      g
.26	       .1957   v
};\label{fig:PFG_AN-R18}
\addplot[
scatter/classes={b=R50,v=R18, g=AN },
scatter,
black,
mark=*, mark size=0.5mm,
scatter src=explicit symbolic,
] table [meta=Class] {
x          y     Class     
.38	      .9560   g
.35	      .0213   b
};\label{fig:DF_AN-R50}
\addplot[
scatter/classes={b=R50,v=R18, g=AN },
scatter,
black,
mark=*, mark size=0.5mm,
scatter src=explicit symbolic,
] table [meta=Class] {
x          y     Class     
.38	       .9560      g
.41	       .0283        v
};\label{fig:DF_AN-R18}
\addplot[
scatter/classes={b=R50,v=R18, g=AN },
scatter,
black,
mark=*, mark size=0.5mm,
scatter src=explicit symbolic,
] table [meta=Class] {
x          y     Class     
.53	       1   g
.50	      .0707   b
};\label{fig:SF_AN-R50}
\addplot[
scatter/classes={b=R50,v=R18, g=AN },
scatter,
black,
mark=*, mark size=0.5mm,
scatter src=explicit symbolic,
] table [meta=Class] {
x          y     Class     
.53	       1      g
.56	       .0660        v
};\label{fig:SF_AN-R18}
\addplot[
scatter/classes={b=R50,v=R18, g=AN },
scatter,
black,
mark=*, mark size=0.5mm,
scatter src=explicit symbolic,
] table [meta=Class] {
x          y     Class     
.68	      .9963   g
.65	      .0180   b
};\label{fig:cw_AN-R50}
\addplot[
scatter/classes={b=R50,v=R18, g=AN },
scatter,
black,
mark=*, mark size=0.5mm,
scatter src=explicit symbolic,
] table [meta=Class] {
x          y     Class     
.68	       .9963      g
.71	       .0274      v
};\label{fig:cw_AN-R18}
\addplot[
scatter/classes={b=R50,v=R18, g=AN },
scatter,
black,
mark=*, mark size=0.5mm,
scatter src=explicit symbolic,
] table [meta=Class] {
x          y     Class     
.83	      .990   g
.80	      .4240   b
};\label{fig:sa_AN-R50}
\addplot[
scatter/classes={b=R50,v=R18, g=AN },
scatter,
black,
mark=*, mark size=0.5mm,
scatter src=explicit symbolic,
] table [meta=Class] {
x          y     Class     
.83	       .990      g
.86	       .4657      v
};\label{fig:sa_AN-R18}
\addplot[
scatter/classes={b=R50,v=R18, g=AN },
scatter,
black,
mark=*, mark size=0.5mm,
scatter src=explicit symbolic,
] table [meta=Class] {
x          y     Class     
.98	      .9893   g
.95	      .3807   b
};\label{fig:ef_AN-R50}
\addplot[
scatter/classes={b=R50,v=R18, g=AN },
scatter,
black,
mark=*, mark size=0.5mm,
scatter src=explicit symbolic,
] table [meta=Class] {
x          y     Class     
.98	       .9893      g
1.01	       .4266      v
};\label{fig:ef_AN-R18}
\end{axis}
\end{tikzpicture}
\begin{tikzpicture}
\node[inner sep=0pt] (whitehead) at (.27,2.6)
{\tiny NF}; 
\node[inner sep=0pt] (whitehead) at (.80,2.6)
{\tiny PF}; 
\node[inner sep=0pt] (whitehead) at (1.30,2.6)
{\tiny DF}; 
\node[inner sep=0pt] (whitehead) at (1.80,2.6)
{\tiny SF};
\node[inner sep=0pt] (whitehead) at (2.30,2.6)
{\tiny CW};  
\node[inner sep=0pt] (whitehead) at (2.77,2.6)
{\tiny SA};  
\node[inner sep=0pt] (whitehead) at (3.30,2.6)
{\tiny \textbf{EF}};    
\begin{axis}[
tiny,
axis lines=left, 
width=5cm,
height=4cm,
cycle list name=color list,
xticklabels={},
xtick=\empty,
xmin=0.00,
ymin=0.00,
ymax=1.01,
xmax=1.01,
smooth,
enlarge x limits=0.01,
x axis line style={draw opacity=0},
y label style={font=\tiny, at={(axis description cs:.2,.5)},anchor=south},
y tick label style={
/pgf/number format/.cd,
fixed,
fixed zerofill,
precision=1,
/tikz/.cd
},
x tick label style={
/pgf/number format/.cd,
fixed,
fixed zerofill,
precision=1,
/tikz/.cd}
]
\addplot[
scatter/classes={b=R50,v=R18, g=AN },
scatter,
black,
mark=*, mark size=0.5mm,
scatter src=explicit symbolic,
] table [meta=Class] {
x          y     Class     
.08	        .9440      g
.05	        .0883   b
};\label{fig:FG_AN-R50}
\addplot[
scatter/classes={b=R50,v=R18, g=AN },
scatter,
black,
mark=*, mark size=0.5mm,
scatter src=explicit symbolic,
] table [meta=Class] {
x          y     Class     
.08	        .9440       g
.11	        .0920   v
};\label{fig:FG_AN-R18}
\addplot[
scatter/classes={b=R50,v=R18, g=AN },
scatter,
black,
mark=*, mark size=0.5mm,
scatter src=explicit symbolic,
] table [meta=Class] {
x          y     Class     
.23	        1      g
.20	       .1270   b
};\label{fig:PFG_AN-R50}
\addplot[
scatter/classes={b=R50,v=R18, g=AN },
scatter,
black,
mark=*, mark size=0.5mm,
scatter src=explicit symbolic,
] table [meta=Class] {
x          y     Class     
.23	        1      g
.26	       .1517   v
};\label{fig:PFG_AN-R18}
\addplot[
scatter/classes={b=R50,v=R18, g=AN },
scatter,
black,
mark=*, mark size=0.5mm,
scatter src=explicit symbolic,
] table [meta=Class] {
x          y     Class     
.38	      .993   g
.35	      .017   b
};\label{fig:DF_AN-R50}
\addplot[
scatter/classes={b=R50,v=R18, g=AN },
scatter,
black,
mark=*, mark size=0.5mm,
scatter src=explicit symbolic,
] table [meta=Class] {
x          y     Class     
.38	       .993      g
.41	       .019        v
};\label{fig:DF_AN-R18}
\addplot[
scatter/classes={b=R50,v=R18, g=AN },
scatter,
black,
mark=*, mark size=0.5mm,
scatter src=explicit symbolic,
] table [meta=Class] {
x          y     Class     
.53	       .9997   g
.50	      .0617   b
};\label{fig:SF_AN-R50}
\addplot[
scatter/classes={b=R50,v=R18, g=AN },
scatter,
black,
mark=*, mark size=0.5mm,
scatter src=explicit symbolic,
] table [meta=Class] {
x          y     Class     
.53	       .9997      g
.56	       .0790        v
};\label{fig:SF_AN-R18}
\addplot[
scatter/classes={b=R50,v=R18, g=AN },
scatter,
black,
mark=*, mark size=0.5mm,
scatter src=explicit symbolic,
] table [meta=Class] {
x          y     Class     
.68	      .965   g
.65	      .0476   b
};\label{fig:cw_AN-R50}
\addplot[
scatter/classes={b=R50,v=R18, g=AN },
scatter,
black,
mark=*, mark size=0.5mm,
scatter src=explicit symbolic,
] table [meta=Class] {
x          y     Class     
.68	       .965      g
.71	       .063      v
};\label{fig:cw_AN-R18}
\addplot[
scatter/classes={b=R50,v=R18, g=AN },
scatter,
black,
mark=*, mark size=0.5mm,
scatter src=explicit symbolic,
] table [meta=Class] {
x          y     Class     
.83	      .9937   g
.80	      .3593   b
};\label{fig:sa_AN-R50}
\addplot[
scatter/classes={b=R50,v=R18, g=AN },
scatter,
black,
mark=*, mark size=0.5mm,
scatter src=explicit symbolic,
] table [meta=Class] {
x          y     Class     
.83	       .9937      g
.86	       .4310      v
};\label{fig:sa_AN-R18}
\addplot[
scatter/classes={b=R50,v=R18, g=AN },
scatter,
black,
mark=*, mark size=0.5mm,
scatter src=explicit symbolic,
] table [meta=Class] {
x          y     Class     
.98	      .9860   g
.95	      .2630   b
};\label{fig:ef_AN-R50}
\addplot[
scatter/classes={b=R50,v=R18, g=AN },
scatter,
black,
mark=*, mark size=0.5mm,
scatter src=explicit symbolic,
] table [meta=Class] {
x          y     Class     
.98	       .9860      g
1.01	       .3157      v
};\label{fig:ef_AN-R18}
\end{axis}
\end{tikzpicture}
\label{fig:acc}
\vspace{-10pt}
\caption{Misleading 
rate and transferability of N-FGSM (NF), P-FGSM (PF), DeepFool (DF), SparseFool (SF), CW, SemanticAdv (SA) and EdgeFool (EF) for ResNet-50~({\protect\tikz \protect\draw[color=R50, fill=R50] plot[mark=*, mark size=0.5mm] (0,0);}), ResNet-18~({\protect\tikz \protect\draw[color=R18, fill=R18] plot[mark=*,mark size=0.5mm] (0,0);}) and AlexNet~({\protect\tikz \protect\draw[color=AN, fill=AN] plot[mark=*, mark size=0.5mm] (0,0);}) on Private-Places365 and ImageNet. EdgeFool is more transferable than other adversarial methods, except SemanticAdv that however severely distorts the colours (see Figure~\ref{fig:advExampleSOA}).}
\label{fig:acc}
\vspace{-10pt}
\end{figure}
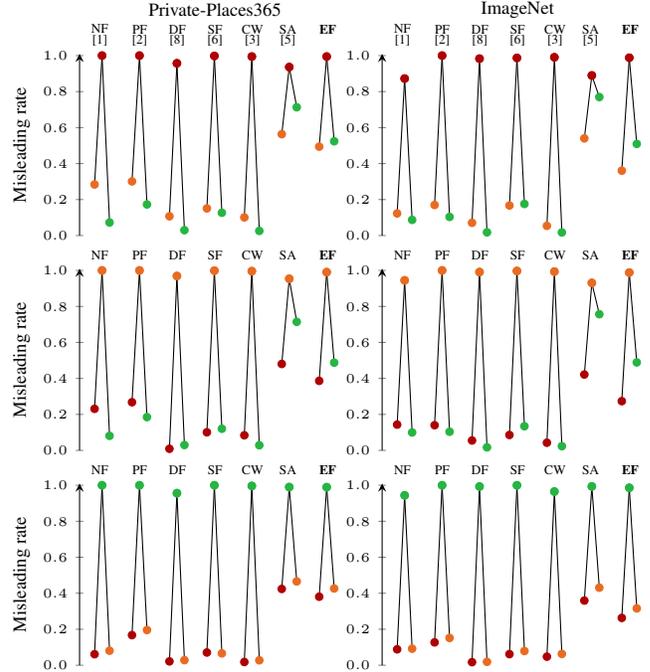

\begin{table*}[t!]
\centering
\caption{Detectability rate~$(\downarrow)$ 
of N-FGSM, P-FGSM, DeepFool, SparseFool, CW, SemanticAdv, and EdgeFool on Private (P)-Places365 and ImageNet for ResNet-50 (R-50), ResNet-18 (R-18) and AlexNet (A) classifiers. For the Quantization (Q), we reduce the number of bits from 8 to 4-bit (4b) to 7-bit (7b). Smoothing (S) is a median filter with $2\times2$ (2m) and $3\times3$ (3m) on the adversarial images. Although SemanticAdv is less detectable than EdgeFool, its adversarial images suffer from unnatural colour changes (see Figure~\ref{fig:advExampleSOA} and Figure~\ref{fig:semAdv}). } 

\vspace{0.5cm}
\setlength{\tabcolsep}{3pt}
\label{tab:D-PC}
\begin{tabular}{|l|c|c|ccc|ccc|ccc|ccc|ccc|ccc|ccc|}
\cline{4-24}
\multicolumn{3}{c|}{ \textbf{}}& \multicolumn{3}{c|}{N-FGSM~\cite{kurakin2016adversarial}} & \multicolumn{3}{c|}{P-FGSM~\cite{Li2019}} & \multicolumn{3}{c|}{DeepFool~\cite{MoosaviDezfooli16}} & \multicolumn{3}{c|}{SparseFool~\cite{modas2018sparsefool}} & \multicolumn{3}{c|}{CW~\cite{carlini2017towards}} & \multicolumn{3}{c|}{SemanticAdv~\cite{hosseini2018semantic}} & \multicolumn{3}{c|}{{EdgeFool}} \\
\cline{4-24}
\multicolumn{3}{c|}{\textbf{}} & R-50 & R-18 & A & R-50 & R-18 & A & R-50 & R-18 & A & R-50 & R-18 & A & R-50 & R-18 & A & R-50 & R-18 & A & R-50 & R-18 & A \\ 
\hline
\multirow{6}{*}{\rotatebox[origin=c]{90}{\textbf{P-Places365}}}
& \multirow{4}{*}{\textbf{Q}}
&\textbf{4b}  & .08 & .07 & .06 & .05 & .02 & .01 & .26 & .23 & .14 & .06 & .05 & .06 & .31 & .24 & .14 & .06 & .06 & .06 & .01 & .00 & .00\\
&&\textbf{5b}  & .05 & .06 & .06 & .01 & .01 & .01 & .27 & .23 & .13 & .07 & .07 & .06 & .33 & .28 & .14 & .09 & .09 & .07 & .01 & .00 & .00\\
&&\textbf{6b}  & .04 & .04 & .06 & .01 & .01 & .01 & .23 & .18 & .15 & .07 & .06 & .06 & .28 & .23 & .13 & .11 & .07 & .07 & .01 & .00 & .00\\
&&\textbf{7b}  & .06 & .05 & .08 & .01 & .01 & .01 & .17 & .15 & .15 & .08 & .06 & .07 & .19 & .15 & .11 & .12 & .09 & .07 & .02 & .02 & .03\\
\cline{2-24}
& \multirow{2}{*}{\textbf{S}}
&\textbf{2m}  & .39 & .36 & .43 & .86 & .82 & .93 & .49 & .49 & .40 & .35 & .34 & .22 & .48 & .45 & .36 & .14 & .11 & .08 & .28 & .27 & .21\\
&&\textbf{3m}  & .39 & .41 & .24 & .96 & .94 & .85 & .35 & .41 & .19 & .17 & .22 & .10 & .33 & .38 & .15 & .12 & .13 & .05 & .19 & .25 & .14\\
\cline{2-24}
\hline\hline
\multirow{6}{*}{\rotatebox[origin=c]{90}{\textbf{ImageNet}}}
& \multirow{4}{*}{\textbf{Q}}
&\textbf{4b}  & .24 & .18 & .11 & .45 & .24 & .02 & .48 & .37 & .22 & .13 & .08 & .07 & .99 & .96 & .68 & .11 & .08 & .08 & .04 & .01 & .01\\
&&\textbf{5b}  & .18 & .16 & .09 & .01 & .01 & .01 & .32 & .25 & .20 & .14 & .11 & .08 & .98 & .97 & .71 & .12 & .11 & .09 & .02 & .01 & .01\\
&&\textbf{6b}  & .12 & .11 & .07 & .01 & .01 & .01 & .20 & .15 & .19 & .14 & .09 & .09 & .97 & .94 & .67 & .14 & .11 & .09 & .03 & .01 & .01 \\
&&\textbf{7b}  & .10 & .09 & .09 & .01 & .01 & .01 & .11 & .09 & .17 & .12 & .11 & .10 & .88 & .85 & .56 & .14 & .12 & .10 & .03 & .03 & .06 \\
\cline{2-24}
& \multirow{2}{*}{\textbf{S}}
&\textbf{2m}  & .40 & .35 & .29 & .95 & .91 & .93 & .62 & .57 & .48 & .42 & .38 & .24 & .99 & .98 & .85 & .14 & .09 & .07 & .32 & .27 & .21\\
&&\textbf{3m}  & .31 & .29 & .17 & .98 & .97 & .93 & .55 & .53 & .24 & .09 & .10 & .11 & .92 & .88 & .51 & .09 & .07 & .05 & .13 & .13 & .12\\
\cline{2-24}

\hline

\end{tabular}
\vspace{10pt}
\end{table*}

\begin{figure*}[t!]
    \centering
    \setlength\tabcolsep{0.2pt}
    \begin{tabular}{cccccccc}
         Orginal & N-FGSM~\cite{kurakin2016adversarial} & P-FGSM~\cite{Li2019} & DeepFool~\cite{MoosaviDezfooli16} & SparseFool~\cite{modas2018sparsefool} & CW~\cite{carlini2017towards} & SemanticAdv~\cite{hosseini2018semantic} & {{EdgeFool}}\\
          \includegraphics[width=0.125\textwidth]{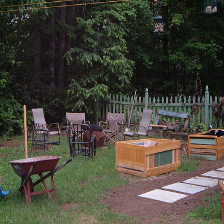}&
           \includegraphics[width=0.125\textwidth]{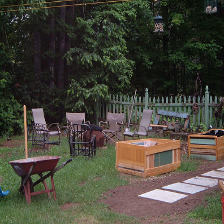}&
           \includegraphics[width=0.125\textwidth]{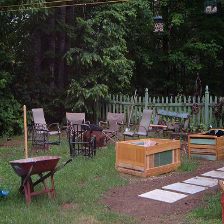}&
           \includegraphics[width=0.125\textwidth]{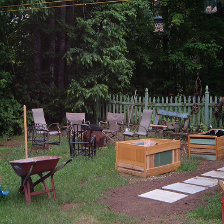}&
           \includegraphics[width=0.125\textwidth]{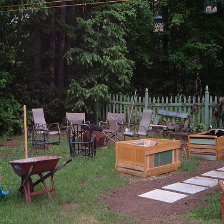}&
           \includegraphics[width=0.125\textwidth]{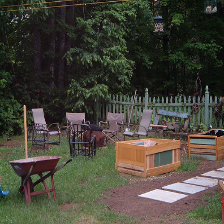}&
           \includegraphics[width=0.125\textwidth]{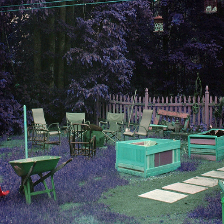}&
           \includegraphics[width=0.125\textwidth]{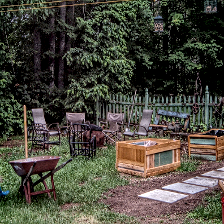}\\
         
         \includegraphics[width=0.125\textwidth]{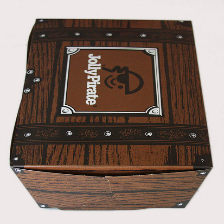}&
         \includegraphics[width=0.125\textwidth]{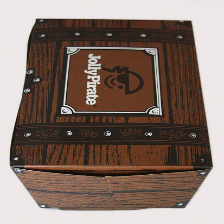}&
         \includegraphics[width=0.125\textwidth]{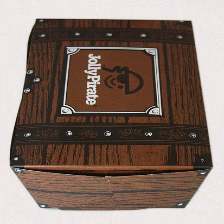}&
         \includegraphics[width=0.125\textwidth]{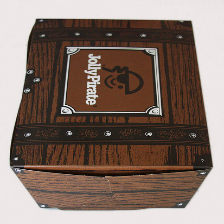}&
         \includegraphics[width=0.125\textwidth]{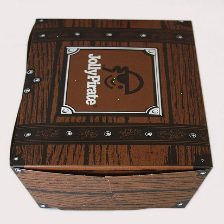}&
         \includegraphics[width=0.125\textwidth]{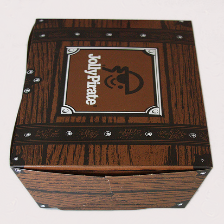}&
         \includegraphics[width=0.125\textwidth]{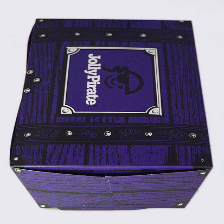}&
         \includegraphics[width=0.125\textwidth]{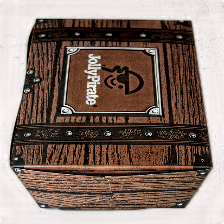}\\

          \includegraphics[width=0.125\textwidth]{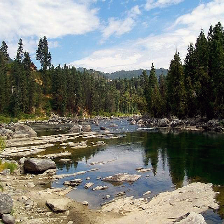}&
           \includegraphics[width=0.125\textwidth]{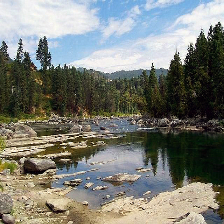}&
           \includegraphics[width=0.125\textwidth]{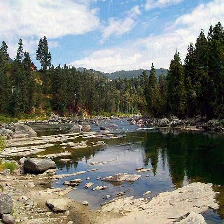}&
           \includegraphics[width=0.125\textwidth]{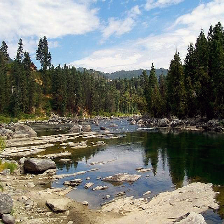}&
           \includegraphics[width=0.125\textwidth]{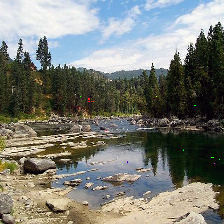}&
          \includegraphics[width=0.125\textwidth]{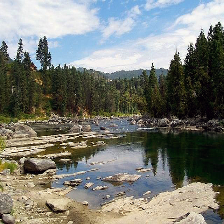}&
           \includegraphics[width=0.125\textwidth]{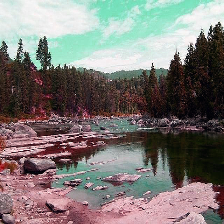}&
           \includegraphics[width=0.125\textwidth]{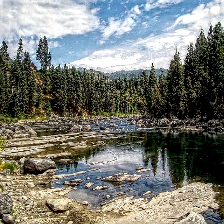}\\
         
    \end{tabular}
      \caption{Adversarial images generated by EdgeFool and other state-of-the-art adversarial methods for ResNet-50 in ImageNet and Private-Places365. N-FGSM, P-FGSM, DeepFool and CW adversarial images are similar to the original images to human eyes, while they are detectable by defence frameworks and not transferable (see Table~\ref{tab:D-PC} and Figure~\ref{fig:acc}). EdgeFool takes advantage of image processing filtering in order to generate structure-aware adversarial perturbations that enhance the details of the image.} 
      \label{fig:advExampleSOA}
      \vspace{10pt}
\end{figure*}

As performance measures we consider the misleading rate, transferability and detectability. The \emph{misleading rate} is the ratio between the number of adversarial images that successfully mislead a classifier and the total number of images. 
\emph{Transferability} is the misleading rate of adversarial images on a classifier that is different from the one that was used to generate the perturbations. Finally,  \emph{detectability} is the ratio between the number of detected adversarial images and the total number of images. To detect an image as adversarial we use the so-called feature squeezing framework~\cite{xu2017feature}, which consists of applying median filtering and bit-depth reduction. An image is considered adversarial if  the $\ell_1$ difference between the $D$-dimentional class predictions of the image, $\dot{\mathbf{p}}$, and its {\em squeezed} version exceed the threshold, which is defined based on the $\ell_1$ difference of the original images and their corresponding squeezed images by accepting a $5\%$ false positive rate.

Figure~\ref{fig:acc} shows the misleading rate and transferability of N-FGSM, P-FGSM, DeepFool, SparseFool, CW, SemanticAdv, and EdgeFool against ResNet-50, ResNet-18, and Alexnet. The top circle and bottom circles indicate the misleading rate against the classifier that adversarial images are generated for and transferability to other classifiers, respectively. 
While most adversarial methods successfully mislead a particular classifier, EdgeFool and SemanticAdv have higher transferability to other classifiers. For example, CW generates adversarial images that mislead ResNet-18 above $99\%$, while only $2\%$ and $4\%$ are transferable to AlexNet and ResNet-50. EdgeFool adversarial images generated for ResNet-50 have above $99\%$ success rate on ResNet-50 and the SemanticAdv misleading rate is $93\%$; however, EdgeFool and SemanticAdv are $52\%$ and $71\%$ transferable to AlexNet. The higher transferability of SemanticAdv is due to the large colour distortions of the generated adversarial images.

\begin{figure}[t!]
    \centering
    \setlength\tabcolsep{1pt}
    \begin{tabular}{ccc}
         \includegraphics[valign=m,width=0.25\linewidth]{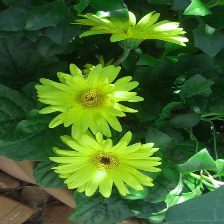}&
         \includegraphics[valign=m,width=0.25\linewidth]{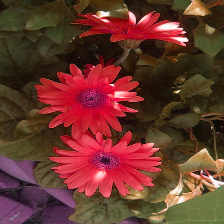}&
         \includegraphics[valign=m,width=0.25\linewidth]{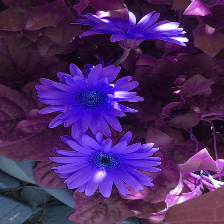}
         \\
          (a) & (b) & (c) \\
    \end{tabular}
    \caption{Examples of adversarial images generated by SemanticAdv~\cite{hosseini2018semantic} against (a) AlexNet, (b) ResNet18 and (c) ResNet50 for the original image in Figure~\ref{fig:advLab}(a). }
      \label{fig:semAdv}
      \vspace{-10pt}
\end{figure}

Table~\ref{tab:D-PC} reports the detectability rate of adversarial images for ResNet-50, ResNet-18 and AlexNet on ImageNet and Private-Places365. 
Median filtering and bit-depth reduction help detect the small but high spatial-frequency perturbations of constrained adversarial methods (especially P-FGSM and CW), while SemanticAdv generates the most undetectable adversarial images to median filtering, as it perturbs hue and saturation of all pixels by the same randomly-generated amount. 
EdgeFool is less detectable than constrained adversarial methods, but more detectable than  SemanticAdv, as some of the enhanced (adversarial) details are smoothed out by the median filtering.
Although EdgeFool and SemanticAdv are comparable in terms of the misleading rate
and transferability, SemanticAdv perturbs each pixel more than EdgeFool and severely distorts the  adversarial images as the colour changes are  random  (see Figure~\ref{fig:advExampleSOA} and Figure~\ref{fig:semAdv}). SparseFool perturbs only a few pixels with large and prominent changes, whereas the adversarial images of
EdgeFool contain  adversarial perturbations, which enhance details.

\section{Conclusion} 
 \vspace{-10pt}
\label{sec:con}
We presented {EdgeFool}, an adversarial image enhancement filter that trains a multi-task fully convolutional neural network to generate adversarial images whose details are enhanced. 
We compared {EdgeFool} with six state-of-the-art adversarial methods on ResNet-50, ResNet-18 and AlexNet classifiers using ImageNet and Private-Places365 datasets and showed that EdgeFool satisfies misleading, transferability and undetectability objectives. As future work, we will validate EdgeFool on other classifiers and with other datasets, as well as perform a formal subjective evaluation of the adversarial image quality.

\end{document}